\documentclass[letterpaper, 10 pt, conference]{ieeeconf}  

\IEEEoverridecommandlockouts                              

\usepackage[nopar]{lipsum}
\usepackage{amsmath,amsfonts}
\usepackage[noend]{algorithmic}
\usepackage{algorithm}
\usepackage{array}
\usepackage{textcomp}
\usepackage{stfloats}
\usepackage{url}
\usepackage{verbatim}
\usepackage{graphicx}
\usepackage{cite}
\usepackage{cleveref}
\usepackage{romannum}

\usepackage{booktabs}
\usepackage{tikz}
\usepackage{tikzscale}
\usepackage{mathrsfs} 
\usepackage{multirow}
\usepackage{wrapfig}
\usepackage{graphicx}
\usepackage[font=footnotesize,labelformat=simple]{subcaption}
\usepackage[font=footnotesize]{caption}
\usepackage{array}
\usepackage{float}
\usepackage{multicol}

\title{\LARGE \bf
A MIP-Based Approach for Multi-Robot \\ Geometric Task-and-Motion Planning
}

\author{Hejia Zhang, Shao-Hung Chan, Jie Zhong, Jiaoyang Li, Sven Koenig, Stefanos Nikolaidis \\
\thanks{Hejia Zhang, Shao-Hung Chan, Jie Zhong, Jiaoyang Li, Sven Koenig and Stefanos Nikolaids are with the Department of Computer Science, University of Southern California, Los Angeles, USA {\tt\small\{hejiazha,shaohung,jzhong54,jiaoyanl,skoenig,\newline nikolaid\}@usc.edu}.}
}

\begin{document}

\maketitle
\thispagestyle{empty}
\pagestyle{empty}

\begin{abstract}

We address multi-robot geometric task-and-motion planning (MR-GTAMP) problems in \textit{synchronous}, \textit{monotone} setups. The goal of the MR-GTAMP problem is to move objects with multiple robots to goal regions in the presence of other movable objects. To perform the tasks successfully and effectively, the robots have to adopt intelligent collaboration strategies, i.e., decide which robot should move which objects to which positions, and perform collaborative actions, such as handovers. To endow robots with these collaboration capabilities, we propose to first collect occlusion and reachability information for each robot as well as information about whether two robots can perform a handover action by calling motion-planning algorithms. We then propose a method that uses the collected information to build a graph structure which captures the precedence of the manipulations of different objects and supports the implementation of a mixed-integer program to guide the search for highly effective collaborative task-and-motion plans. The search process for collaborative task-and-motion plans is based on a Monte-Carlo Tree Search (MCTS) exploration strategy to achieve exploration-exploitation balance. We evaluate our framework in two challenging GTAMP domains and show that it can generate high-quality task-and-motion plans with respect to the planning time, the resulting plan length and the number of objects moved compared to two state-of-the-art baselines.

\end{abstract}

\section{INTRODUCTION}

Task-and-motion planning (TAMP) is the problem of combining task and motion planning to divide an objective, such as assembling a table, into a series of robot-executable motion trajectories~\cite{doi:10.1146/annurev-control-091420-084139}. Task planning is used to generate a sequence of discrete actions, such as pick up a screwdriver and drive a screw, while motion planning is used to compute the actual trajectories the robot should execute.

Geometric task-and-motion planning (GTAMP) is an important subclass of TAMP where the robot has to move several objects to regions in the presence of other movable objects~\cite{Kim2019}. Previously, GTAMP has been addressed efficiently in single-robot domains~\cite{Kim2019,pmlr-v100-kim20a,doi:10.1177/02783649211038280}. We focus on \textit{multi-robot geometric task-and-motion planning} (MR-GTAMP), where the robots have to collaborate to move several objects to regions in the presence of movable obstacles.

\begin{figure}[!t]
\centering
\includegraphics[width=0.9\linewidth]{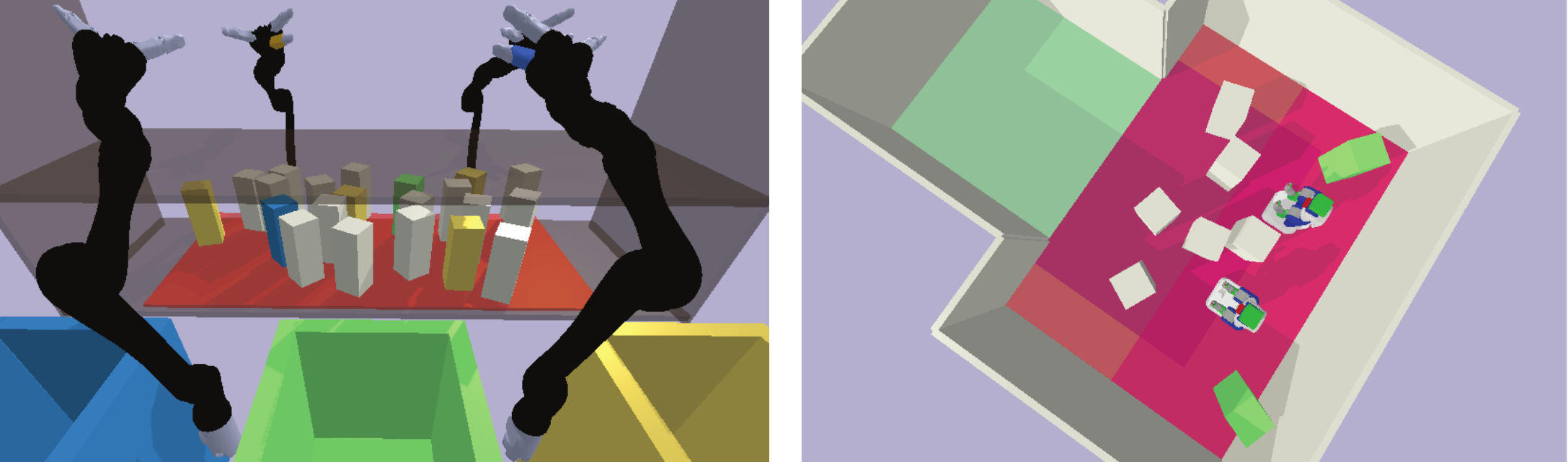}
\caption{Left: Packing colored objects into boxes. Right: Moving the colored boxes to the green region.}
\label{fig:best}
\end{figure}

MR-GTAMP naturally arises in many multi-robot manipulation domains, such as multi-robot construction, assembly and autonomous warehousing~\cite{chen2022cooperative,hartmann2021long}. MR-GTAMP is interesting as multi-robot systems can perform manipulation tasks more effectively than single-robot systems and can also perform manipulation tasks that are beyond the capabilities of single-robot systems~\cite{9357998}. For example, in a product-packaging task, a single robot may have to move a lot of objects to clear a path to grasp an object, while a two-robot system can easily perform a handover action to increase the effectiveness of task execution. Examples of MR-GTAMP problems are shown in Figure~\ref{fig:best}.

We address the following research question: How can we generate collaboration strategies for multiple robots to perform GTAMP tasks effectively?


Determining effective collaborative action sequences for multiple robots is difficult as manipulation planning among movable obstacles has been shown to be NP-hard in the single-robot domain~\cite{4209604,9197485}. MR-GTAMP is even harder since one needs to decide which robot should move which objects to which positions.

Our key insight to solving MR-GTAMP efficiently is that \textit{we can compute information about the manipulation capabilities of individual robots and their potential collaborative relationships by calling motion-planning algorithms} to prune the search space and guide the search process. For example, based on the information that a robot cannot reach an object, we can eliminate all task plans that involve the action where the robot has to reach the object. Moreover, the computed information can be used to generate collaborative plans where each robot can perform the tasks that it excels at.

\begin{figure}[!t]
\centering
\includegraphics[width=1.\linewidth]{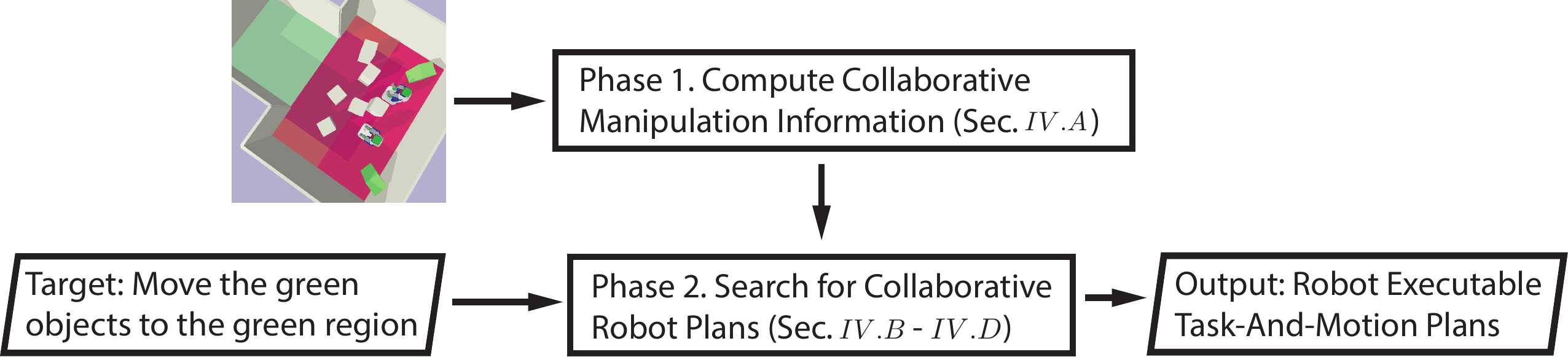}
\caption{Overview of the proposed framework.}
\label{fig:summary}
\end{figure}

  
We propose a two-phase framework. In the first phase, we compute the collaborative manipulation information, i.e., the occlusion and reachability information for individual robots and the potential collaborative relationships between them (Sec.~\ref{sec:rep}). In the second phase, we search for collaborative task-and-motion plans using a Monte-Carlo Tree Search (MCTS) exploration strategy due to its good exploration-exploitation balance (Sec.~\ref{sec:tree_search}). Our search algorithm is based on two key components: (\romannum{1}) the first key component generates promising task skeletons for moving a specified set of objects with the collected information from the first phase by formulating a series of mixed-integer linear programs (MIPs), that can be solved efficiently by leveraging recent developments in MIP solvers~\cite{cplex2009v12} (Sec.~\ref{sec:approx}); and (\romannum{2}) the second key component efficiently finds feasible continuous parameters for the generated task skeletons, such as the locations to which to relocate objects (Sec.~\ref{sec:inst}). Fig.~\ref{fig:summary} presents an overview of our framework.

We compare our framework with two state-of-the-art baselines, namely, a general MR-TAMP framework~\cite{9636119} and a multi-robot extension of the ResolveSpatialConstraints (RSC) algorithm~\cite{4209604}. We show that our framework can solve MR-GTAMP problem instances that are challenging for the baseline methods. We also show that our framework can generate high-quality task-and-motion plans with respect to the planning time, the resulting plan length and the number of objects moved compared to the baselines (Sec.~\ref{sec:exp}).



Our work makes the following assumptions, which are common in MR-TAMP~\cite{9357998,9636119}: (\romannum{1}) it considers only \textit{monotone} instances of the MR-GTAMP problem, where each object is moved only once; and (\romannum{2}) the robots \textit{synchronously} start and stop the executions of actions. We plan to relax these assumptions in future work.

\section{RELATED WORK}

There has been much work on solving single-robot GTAMP (SR-GTAMP) problems efficiently~\cite{doi:10.1177/02783649211038280,Kim2019,pmlr-v100-kim20a} by utilizing learning to guide planning. Several existing problem types in the literature can also be seen as GTAMP problems. In~\cite{4209604}, the ``manipulation among movable obstacles" (MAMO) problem is addressed, in which a robot moves objects out of the way to move a specified object to its goal location. In~\cite{9197485} and~\cite{9196652,8794143}, the object retrieval problem is addressed, in which a target object has to be retrieved from clutter by relocating the surrounding objects. In~\cite{7487583,7487581}, the rearrangement planning problem is addressed, in which a robot is tasked to move objects into a given configuration. However, these methods do not plan collaboration strategies in multi-robot domains.


There has been work in multi-robot domains on general task and motion planning~\cite{9636119,7989464,10.3389/frobt.2021.637888}. We focus on a subclass of these problems, where we wish to move objects in the presence of movable obstacles. In~\cite{7759624,ahn2021coordination}, efficient approaches are proposed for the multi-robot object retrieval problem, assuming permanent object removal and considering one target object at a time, while our planner considers several target objects at the same time and relocates the obstacles within the workspace. Multi-robot rearrangement planning problems~\cite{9357998,hartmann2021long,chen2022cooperative} are also closely related to MR-GTAMP. However, the rearrangement planning problems assume that the goal configurations of the objects are given, while MR-GTAMP requires the planners to decide which objects to move and to which positions. There is also work that focuses on task allocation and scheduling for multiple robots, assuming that a sequence of discrete actions to be executed is given~\cite{9197103}. However, MR-GTAMP requires the planners to decide which discrete actions to execute, e.g., which objects to move.



\section{PROBLEM FORMULATION}\label{sec:formulation}
In a MR-GTAMP problem, we have a set of $n_\mathbf{R}$ robots $\mathbf{R} = \{R_i\}^{n_{\mathbf{R}}}_{i=1}$, a set of fixed rigid objects $\mathbf{F}$, a set of $n_\mathbf{M}$ movable rigid objects $\mathbf{M} = \{M_i\}^{n_\mathbf{M}}_{i=1}$ and a set of $n_\mathbf{Re}$ regions $\mathbf{Re} = \{{Re}_i\}^{n_{\mathbf{Re}}}_{i=1}$. We assume that all objects and regions have known and fixed shapes. The focus of our work is not on grasp planning~\cite{7743537}. So, for simplicity, we assume a fixed set of grasps $\mathbf{Gr}_{M, R}$ for each object $M \in \mathbf{M}$ and robot $R \in \mathbf{R}$ pair. We denote the union of the sets of grasps for all object and robot pairs as $\mathbf{Gr}$.

Each object has a configuration, which includes its position and orientation. Each robot has a configuration defined in its base pose space and joint space. We are given the initial configurations of all robots, objects and regions and a goal specification $\mathcal{G}$ in form of a conjunction of statements of the form $\textsc{InRegion}(M, Re)$, where $M \in \mathbf{M}$ and $Re \in \mathbf{Re}$, which is true \textit{iff} object $M$ is contained entirely in region $Re$.

We define a grounded joint action as a set of $n_{\mathbf{R}}$ actions and motions performed by each robot at one time step, i.e., the grounded joint action at time step $j$ is an $n_\mathbf{R}$-tuple $s_j = \langle (a_{R_1}^{j}, \xi_{R_1}^{j}), (a_{R_2}^j, \xi_{R_2}^{j}), \dots, (a^j_{R_{n_{\mathbf{R}}}}, \xi_{R_{n_{\mathbf{R}}}}^{j}) \rangle$, where each action $a$ is a pick-and-place action or a wait\footnote{As in~\cite{9636119}, a robot with a wait action does not have to do anything but can move to avoid other robots.} action that the corresponding robot executes and motion $\xi$ is a trajectory that the corresponding robot executes, specified as a sequence of robot configurations. In this work, we focus on pick-and-place actions because of their importance in robotic manipulation in cluttered space. Each pick-and-place action is a tuple composed of $\langle M, Re, R^{pick}, R^{place}, g^{pick}, g^{place}, P_M^{place} \rangle$, where $M$ represents the object to move; $Re$ represents the target region for $M$; $R^{pick}$ and $R^{place}$ represent the robots that pick and place $M$, respectively; $g^{pick}$ and $g^{place}$ represent the grasps used by $R^{pick}$ and $R^{place}$, respectively, and $P_M^{place}$ represents the pose at which to place $M$.  Moreover, we call a pick-and-place action whose $R^{pick}$ is different from $R^{place}$ a handover action. Each grounded joint action will map the configurations of the movable objects to new configurations where the moved objects are at their new poses and the unaffected objects remain at their old poses.

We define a partially grounded joint action as an $n_{\mathbf{R}}$-tuple of the form $\langle \bar{a}_{R_1}, \dots, \bar{a}_{R_{n_{\mathbf{R}}}} \rangle$, where $\bar{a}$ is a wait action or a pick-and-place action without the placement information $P_M^{place}$. We refer to a pick-and-place action without the placement information as a partially grounded pick-and-place action since it has only the information about the grasps that will be used.

We define a task skeleton $\bar{\mathbf{S}}$ as a sequence of partially grounded joint actions. We want to find a task-and-motion plan, i.e., a sequence of grounded joint actions $\mathbf{S}$ to change the configurations of the objects to satisfy $\mathcal{G}$.

A task-and-motion plan, is valid \textit{iff}, at each time step $j$: (\romannum{1}) the corresponding multi-robot trajectory $\Xi^j = \langle \xi_{R_1}^{j}, \xi_{R_2}^{j}, \dots, \xi_{R_{n_{\mathbf{R}}}}^{j} \rangle$ is collision-free; (\romannum{2}) the robots can use the corresponding motion trajectories and grasp poses to grasp the target objects and place them at their target poses without collisions; and (\romannum{3}) all handover actions can be performed without inducing collisions. The considered collisions include collisions between robots, collisions between an object and a robot, and collisions between objects. 
 

\section{OUR APPROACH}
We present our two-phase MR-GTAMP framework (Fig.~\ref{fig:summary}) in this section. In the first phase, we compute the collaborative manipulation information, i.e., the occlusion and reachability information for individual robots and the potential collaborative relationships between the robots (Sec.~\ref{sec:rep}). In the second phase, we use a Monte-Carlo Tree Search exploration strategy to search for task-and-motion plans (Sec.~\ref{sec:tree_search}). The search process depends on a key component that generates promising task skeletons (Sec.~\ref{sec:approx}) and a key component that finds feasible object placements and motion trajectories for the task skeletons to construct executable task-and-motion plans (Sec.~\ref{sec:inst}).

\subsection{Computing Collaborative Manipulation Information}\label{sec:rep}

Given a MR-GTAMP problem instance and the initial configurations of all objects and robots, our framework first computes the occlusion and reachability information for individual robots, e.g., whether an object blocks a robot from manipulating another object and whether a robot can reach a region to place an object there. We also compute whether two robots can perform a handover action for an object by computing whether they can both reach a predefined handover point to transfer the object. In this work, we only consider handover actions for objects that are named in goal specification $\mathcal{G}$ for computational simplicity. We assume that all robots will return to their initial configurations after each time step. Inspired by~\cite{doi:10.1177/02783649211038280}, we use a conjunction of all true instances of a set of predicates to represent the computed information. To define these predicates, we need to define two volumes of workspace similar to~\cite{4209604,doi:10.1177/02783649211038280}. The first volume $V_{pick}(M, g, R, \xi)$ is the volume swept by robot $R$ to grasp object $M$ with grasp $g$ following trajectory $\xi$. The second volume $V_{place}(M, g, R, P^{place}_M, \xi)$ is the volume swept by robot $R$ and object $M$ to transfer the object to pose $P^{place}_M$ after trajectory $\xi$. Our predicates are as follows:

{
\begin{itemize}
    \item \textsc{OccludesPick}$(M_1, M_2, g, R)$ is true \textit{iff} object $M_1$  overlaps with the swept volume $V_{pick}(M_2, g, R, \xi)$, where $\xi$ is chosen to be collision-free, if possible;
    \item \textsc{OccludesGoalPlace}$(M_1, M_2, Re, g, R)$ is true \textit{iff} $M_1$ is an object that overlaps with the swept volume $V_{place}(M_2, g, R, P_{M_2}^{place}, \xi)$, where $P_{M_2}^{place}$ and $\xi$ are chosen to be collision-free, if possible, and the pair $\langle M_2, Re \rangle$ is named in goal specification $\mathcal{G}$;
    \item \textsc{ReachablePick}$(M, g, R)$ is true \textit{iff} there exists a trajectory for robot $R$ to pick object $M$ with grasp $g$;
    \item \textsc{ReachablePlace}$(M, Re, g, R)$ is true \textit{iff} there exists a trajectory for robot $R$ to place object $M$ in region $Re$ with grasp $g$; and 
    \item\textsc{EnableGoalHandover}$(M, g_1, g_2, R_1, R_2)$ is true \textit{iff} two robots $R_1$ and $R_2$ can both reach a predefined handover point for object $M$ with grasps $g_1$ and $g_2$, respectively, and the object $M$ is named in goal specification $\mathcal{G}$. 
\end{itemize}
}

For a predicate instance to be true, the corresponding trajectories are required to be collision-free with respect to the given fixed objects. For a predicate instance of \textsc{EnableGoalHandover} to be true, the two robots should not collide with each other.

The values of all the predicate instances can be computed with existing inverse-kinematics solvers~\cite{diankov_thesis} and motion planners~\cite{lavalle2006planning}. Ideally, we wish to find trajectories for the robots that have the minimum number of collisions with the given objects, i.e., the \textit{minimum constraint removal}~\cite{Hauser-RSS-13} trajectories. However, this is known to be very costly. Thus, we follow previous work~\cite{doi:10.1177/02783649211038280} and first attempt to find a collision-free trajectory with respect to the movable and fixed objects. If we fail, we attempt to find a collision-free trajectory with respect to only the fixed objects. 

In our implementation, we efficiently compute the predicates for individual robots -- with the exception of \textsc{EnableGoalHandover} -- in parallel by creating an identical simulation environment for each robot.

\subsection{Searching for Task-and-Motion Plans}\label{sec:tree_search}

\begin{figure}[!t]
\centering
\includegraphics[width=1.\linewidth]{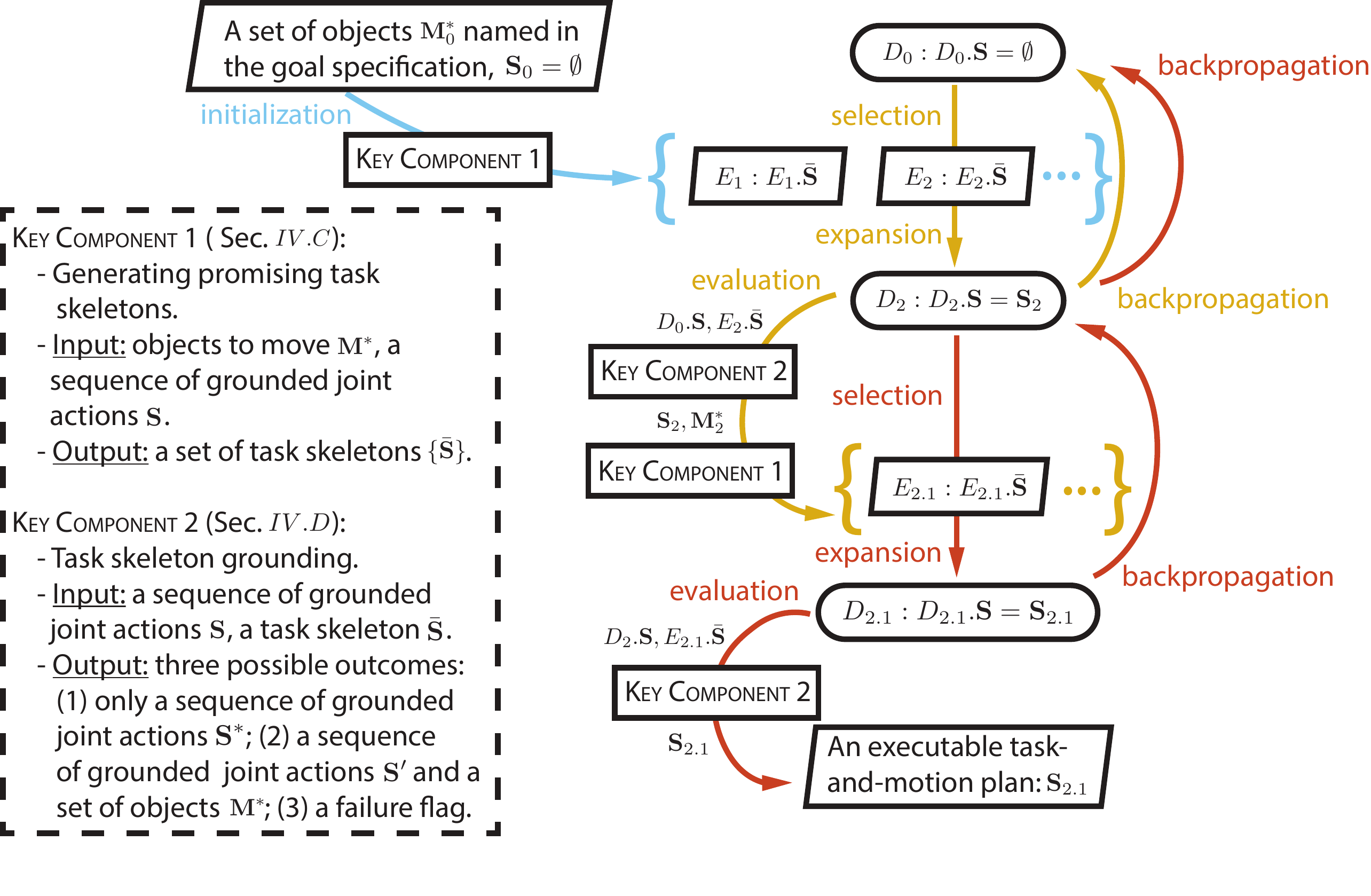}
\caption{Summary of the search process in the second phase of our framework. Blue arrows represent the workflow for initializing the search tree. Yellow arrows represent a search iteration that results in an updated set of objects to be moved and thus a new set of task skeletons to be grounded. Red arrows represent a search iteration that results in an executable plan.}
\label{fig:tree_search}
\end{figure}

We now describe our search process (Fig.~\ref{fig:tree_search}) for efficiently finding high-quality collaborative task-and-motion plans. Our search process generates a search tree whose nodes, denoted as $D$, store sequences of grounded joint actions, denoted as $D.\mathbf{S}$, and whose edges, denoted as $E$, store task skeletons, denoted as $E.\bar{\mathbf{S}}$. At each search iteration, we will select an task skeleton to ground. We define a reward function, which will be described in details later, as the optimization target for task-skeleton selection. The value of an edge is the cumulated reward it has received since the search starts.

Assume that we have a node $D_j$ and an edge $E_i$ coming out of node $D_j$. If we successfully ground task skeleton $E_i.\bar{\mathbf{S}}$, given a sequence of already grounded joint actions $D.\mathbf{S}$, with the task-skeleton grounding component (Sec.~\ref{sec:inst}), the resulting executable task-and-motion plan is a sequence of grounded joint actions with $D.\mathbf{S}$ as postfix. However, there can be situations, where a task skeleton cannot be grounded without moving some objects that are not planned to be moved in that task skeleton (Sec.~\ref{sec:inst}). In these situations, we generate new task skeletons to move those objects with the task-skeleton generating component (Sec.~\ref{sec:approx}).





We propose a Monte-Carlo Tree Search (MCTS) exploration strategy to balance exploration (exploring different candidate task skeletons) and exploitation (biasing the search towards the branches that have received high rewards).


We first generate an initial set of task skeletons (Sec.~\ref{sec:approx}) for moving a set of objects named in the goal specification, utilizing the computed collaborative manipulation information (Sec.~\ref{sec:rep}). We then initialize the search tree by adding a root node $D_0$ for selecting from the initial set of task skeletons. At each search iteration, we have four phases: \textit{selection}, \textit{expansion}, \textit{evaluation} and \textit{backpropagation}. 

\noindent\textbf{Notation.} We use $\lvert \mathbf{S}\rvert$ and $\lvert \bar{\mathbf{S}}\rvert$ to denote the number of objects intended to be moved in sequences of grounded joint actions $\mathbf{S}$ and task skeletons $\bar{\mathbf{S}}$, respectively.

\noindent\textbf{\textit{Selection} phase.} In the \textit{selection} phase, we start at the root node and recursively select the edge with the highest Upper Confidence Bound (UCB) value until we reach an edge $E_i$ with a task skeleton that has not been grounded yet. We denote the tail node of edge $E_i$ as $D_j$. We follow the UCB value formula used in~\cite{silver2017mastering}. The UCB value of the pair of node $D_j$ and edge $E_i$ is: $Q(D_j, E_i) = \frac{E_i.value}{E_i.visits + 1} + c \times E_i.prior \times \frac{\sqrt{D_{j}.visits}}{E_i.visits + 1}$, where $E_i.value$ is the cumulated reward edge $E_i$ has received so far, $D_j.visits$ and $E_i.visits$ are the number of times $D_j$ and $E_i$ have been selected, $c$ is a constant to balance exploration and exploitation, and $E_i.prior$ is used to bias the search with domain knowledge~\cite{silver2017mastering}. In our implementation, we set $E_{i}.prior$ to $\frac{1}{\lvert E_i.\bar{\mathbf{S}}\rvert}$ to prioritize grounding the task skeletons with fewer objects to move. The value of an edge $E_i.value$ is initialized to $0$.


Assume that we select edge $E_i$ at node $D_j$ in the \textit{selection} phase. 

\noindent\textbf{\textit{Expansion} phase.} In the \textit{expansion} phase, we create a new node $D_{j.i}$ as the head node of edge $E_i$.

\noindent\textbf{\textit{Evaluation} phase.} In the \textit{evaluation} phase, we use the task-skeleton grounding component (Sec.~\ref{sec:inst}) to ground task skeleton $E_i.\bar{\mathbf{S}}$ associated with $E_i$ to compute reward $r$ for selecting edge $E_i$. There are three possible outcomes: (\romannum{1}) If we fail at grounding, we set $r$ to $0$. (\romannum{2}) If we obtain a sequence of grounded joint actions $\mathbf{S}^*$, then we found an executable task-and-motion plan. In this case, we set $r$ to $1 + \alpha \frac{1}{\lvert \mathbf{S}^*\rvert}$, where $\alpha$ is a constant hyperparameter used to balance the two terms of the reward and is set to $1$ in our experiments (Sec.~\ref{sec:exp}). The first term of the reward motivates the search algorithm to select branches where more actions have been grounded, and the second term motivates the search algorithm to select branches that will move fewer objects. (\romannum{3}) If we obtain a sequence of grounded joint actions $\mathbf{S}'$ and a set of objects $\mathbf{M}^*$, then we have to move objects $\mathbf{M}^*$ to transport the already grounded joint actions $\mathbf{S}'$ into an executable task-and-motion plan. In this case, we call the task-skeleton generating component (Sec.~\ref{sec:approx}) to move $\mathbf{M}^*$. If we can not find any task skeleton to move $\mathbf{M}^*$, then we set $r$ to $0$. However, if we find a set of task skeletons $\{\bar{\mathbf{S}}\}$, then we set $r$ to $\frac{\mathbf{S}'.length}{\mathbf{S}'.length + \bar{\mathbf{S}}^*.length} + \alpha \frac{1}{\lvert \mathbf{S}'\rvert + \lvert \bar{\mathbf{S}}^* \rvert}$, where $\bar{\mathbf{S}}^*$ is the task skeleton with the minimum number of time steps in $\{\bar{\mathbf{S}}\}$, $\mathbf{S}'.length$ and $\bar{\mathbf{S}}^*.length$ represent the number of time steps of $\mathbf{S}'$ and the number of time steps of $\bar{\mathbf{S}}^*$, respectively.

We use node $D_{j.i}$ to store the returned grounded joint actions $\mathbf{S}'$ as $D_{j.i}.\mathbf{S}$. In the third scenario, if we find new task skeletons we create new edges to store them for node $D_{j.i}$. If no new edge is created, we mark node $D_{j.i}$ as a terminal node.

\noindent\textbf{\textit{Backpropagation} phase.} In the \textit{backpropagation} phase, we update the cumulated reward of the selected edges $\{E^{sel}\}$ with the computed reward $r$ according to $E^{sel}.value = {E^{sel}.value + r}$. We also increment the number of visits of the selected edges and nodes by $1$.

In our implementation, we keep tracking the grounding failures for different task skeletons similar to~\cite{ren2021extended}, so that we can efficiently skip over those branches where grounding their task skeletons is known to be infeasible.



\subsection{Key Component 1: Generating Promising Task Skeletons}\label{sec:approx}

One key component in the second phase (Sec.~\ref{sec:tree_search}) of our framework is to generate promising task skeletons $\{\bar{\mathbf{S}}\}$, i.e., sequences of actions without the placement and trajectory information, for moving a set of objects $\mathbf{M}^*$ given a sequence of already grounded joint actions $\mathbf{S}'$. It will be called at the initialization stage of the search process, where $\mathbf{S}'$ is empty and $\mathbf{M}^*$ is the set of objects named in the goal specification of the problem instance. It will also be called during the search process when the third scenario happens in the \textit{evaluation} phase. The task-skeleton generating algorithm is designed in a way such that we can utilize the computed collaborative manipulation information from the first phase (Sec.~\ref{sec:rep}) to eliminate task plans that include infeasible actions and prioritize motion planning for high-quality task plans that have a small number of time steps and a small number of objects to be moved.

\noindent\textbf{Notation.} Assume that we want to generate task skeletons to move objects $\mathbf{M}^*$ given a sequence of grounded joint actions $\mathbf{S}'$. The set of objects included in $\mathbf{S}'$ cannot be moved again because of the \textit{monotone} assumption. For simplicity of presentation, we slightly abuse $\mathbf{M}$ to denote the movable objects not included in $\mathbf{S}'$.

\begin{figure}[!t]
\centering
\includegraphics[width=0.8\linewidth]{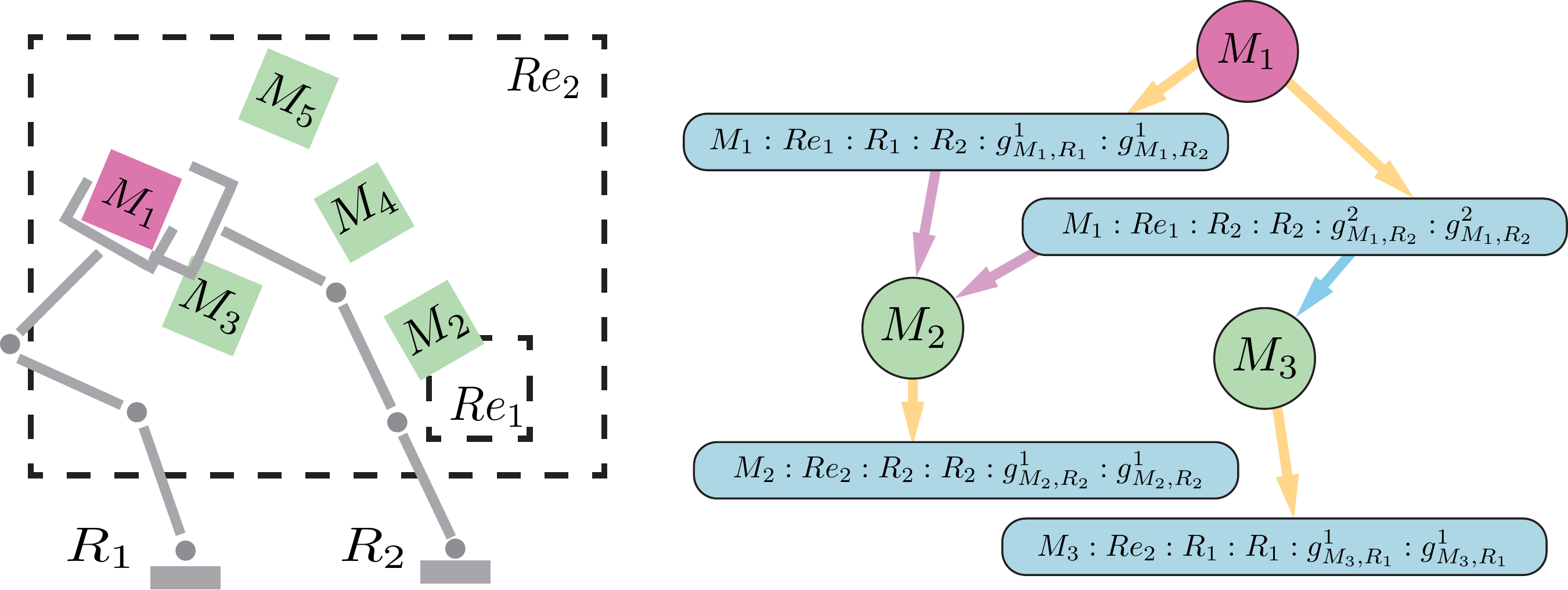}
\caption{(Left) An example scenario where we want to generate task skeletons to move object $\mathbf{M}_1$ given an empty sequence of grounded joint actions. (Right) The corresponding \textit{collaborative manipulation task graph} for moving object $M_1$. The rounded rectangular nodes are \textit{action nodes}. The circular nodes are \textit{object nodes}. The red circular nodes represent objects that are specified to be moved. The yellow arrows represent \textit{action edges}. The purple arrows represent \textit{block-place edges}, and the blue arrow represents a \textit{block-pick edge}.}
\label{fig:mg}
\end{figure}

\noindent\textbf{Building the collaborative manipulation task graph.} To reason about the collaborative manipulation capabilities of the individual robots, we encode the computed information as a graph. We build a \textit{collaborative manipulation task graph} (CMTG) to capture the precedence of the manipulations of different objects, i.e., we can only move an object after we move the obstacles that block the pick-and-place action we are going to execute, based on the computed information from the first phase (Sec.~\ref{sec:rep}). Since we only compute occlusion information for placing objects named in the goal specification, the precedences encoded in the CMTG lack occlusion information for relocating objects that are not named in the goal specification. Instead, we assume that we will always find the feasible places to relocate these objects. We determine the exact object placements during task-skeleton grounding (Sec.~\ref{sec:inst}).

A CMTG (Fig.~\ref{fig:mg}) has two types of nodes: An \textit{object node} represents an object $M \in \mathbf{M}$; and an \textit{action node} represents a partially grounded pick-and-place action $\bar{a}$, i.e. a pick-and-place action without placement information.  A CMTG has three types of edges: An \textit{action edge} is an edge from an object node to an action node. It represents moving the object represented by the object node with the action represented by the action node. A \textit{block-pick edge} is an edge from an action node to an object node. It represents that the object represented by the object node obstructs the pick action of the action represented by the action node. A \textit{block-place edge} is an edge from an action node to an object node. It represents that the object represented by the object node obstructs the place action of the action represented by the action node. All \textit{block-place edges} are connected to the action nodes that move the objects named in the goal specification. A CMTG has a set of object nodes that represents the input objects $\mathbf{M}^*$ that must be moved.


Given the computed collaborative manipulation information and a set of objects $\mathbf{M}^{*}$ to move, we incrementally construct a CMTG by iteratively adding object $M \in \mathbf{M}^*$ to the CMTG with Alg.~\ref{alg:add_movable}. Given the CMTG $\mathbf{C}$ built so far and an object $M$ to add, we first add an object node representing $M$ to $\mathbf{C}$ (Alg.~\ref{alg:add_movable}, line~\ref{line:add_object_node}). Then, for each pair of a robot $R \in \mathbf{R}$ and its grasp $g_{M, R} \in \mathbf{Gr}_{M, R}$, we find all partially grounded pick-and-place actions $\mathbf{\bar{a}}$ that move object $M$ to its target region $Re_M$ with $R$ as the pick robot (Alg.~\ref{alg:add_movable}, line~\ref{line:get_goal_region}-\ref{line:get_actions}). For each partially grounded pick-and-place action $\bar{a}$, we find all movable objects that block the pick action of $\bar{a}$ and add the corresponding block-pick edges (Alg.~\ref{alg:add_movable}, line~\ref{line:pick_block}-\ref{line:pick_block_end}). If $M$ is named in goal specification $\mathcal{G}$, then we also find all movable objects that block the place action of $\bar{a}$ and add the corresponding block-place edges (Alg.~\ref{alg:add_movable}, line~\ref{line:place_block}-\ref{line:place_block_end}). We recursively add the blocking objects in a similar way (Alg.~\ref{alg:add_movable}, lines~\ref{line:explore_pick} and~\ref{line:explore_place}).

\begin{algorithm}[!t]
\caption{\textsc{AddObject}($M, \mathbf{C}$)}\label{alg:add_movable}
 \algsetup{linenosize=\tiny}
  \scriptsize
\begin{algorithmic}[1]
\IF {$M \in \mathbf{C}.object\_nodes$}
\RETURN
\ENDIF
\STATE {$\mathbf{C}.object\_nodes.add(M)$} \label{line:add_object_node}
\IF {$M$ is named in goal specification $\mathcal{G}$}\label{line:get_goal_region}
\STATE {$Re_M$ = \textsc{GetGoalRegion}($M$)}
\ELSE 
\STATE {$Re_M$ = \textsc{GetCurrentRegion}($M$)}
\ENDIF
\FOR {$R^{pick} \in \mathbf{R}$}
    \FOR {$g_{M, R^{pick}} \in \mathbf{Gr}_{M, R^{pick}}$}
        \STATE{$\mathbf{\bar{a}} = \{\}$}
        \IF {\textsc{ReachablePick}$(M, g_{M, R^{pick}}, R^{pick})$}
             \IF {\textsc{ReachablePlace}\\~~~~$(M, Re_M, g_{M, R^{pick}}, R^{pick})$}
                \STATE{$\mathbf{\bar{a}}.add((M, Re_M, R^{pick},R^{pick},$ \\ ~~~~~~~~~$g_{M, R^{pick}},g_{M, R^{pick}}))$}
             \ENDIF
            \IF {$M$ is named in goal specification $\mathcal{G}$}
                \FOR {$R^{place} \in \mathbf{R} \setminus \{R^{pick}\}$}
                    \FOR {$g_{M, R^{place}} \in \mathbf{Gr}_{M, R^{place}}$}
                        \IF {\textsc{EnableGoalHandover}\\$(M, g_{M, R^{pick}}, g_{M, R^{place}}, R^{pick}, R^{place})$ and \\ \textsc{ReachablePlace}\\$(M, Re_M, g_{M, R^{place}}, R^{place})$}
                            \STATE {$\mathbf{\bar{a}}$.add($(M, Re_M, R^{pick},R^{place},$ \\ ~~~~~~~~$g_{M, R^{pick}},g_{M, R^{place}})$)}
                        \ENDIF
                    \ENDFOR 
                \ENDFOR
            \ENDIF\label{line:get_actions}
            \FOR {$\bar{a} \in \mathbf{\bar{a}}$}
                \STATE{$R^{pick}_{\bar{a}}$ is the robot to pick $M$ in $\bar{a}$}
                \STATE{$g^{pick}_{\bar{a}}$ is the grasp used by $R^{pick}_{\bar{a}}$ in $\bar{a}$}
                \STATE{$R^{place}_{\bar{a}}$ is the robot to place $M$ in $\bar{a}$}
                \STATE{$g^{place}_{\bar{a}}$ is the grasp used by $R^{place}_{\bar{a}}$ in $\bar{a}$}
                
                \STATE{$\mathbf{C}.action\_nodes.add(\bar{a})$}
                \STATE {$\mathbf{C}.action\_edges.add(M, \bar{a})$}
                
                \FOR {$M_j \in \mathbf{M}$}\label{line:pick_block}
                    \IF {\textsc{OccludesPick}\\~~~~$(M_j, M, g^{pick}_{\bar{a}}, R^{pick}_{\bar{a}})$}
                        \STATE {\textsc{AddObject}($M_j, \mathbf{C}$)}\label{line:explore_pick}
                        \STATE {$\mathbf{C}.block\_pick\_edges.add(\bar{a}, M_j)$}
                    \ENDIF
                \ENDFOR\label{line:pick_block_end}

                \IF{$M$ is named in goal specification $\mathcal{G}$}\label{line:place_block}
                \FOR {$M_j \in \mathbf{M}$}
                    \IF{\textsc{OccludesGoalPlace}\\~$(M_j, M, Re_M, g^{place}_{\bar{a}}, R^{place}_{\bar{a}})$}
                        \STATE {\textsc{AddObject}($M_j, \mathbf{C}$)}\label{line:explore_place}
                        \STATE {$\mathbf{C}.block\_place\_edges.add(\bar{a}, M_j)$}
                    \ENDIF 
                \ENDFOR 
                \ENDIF\label{line:place_block_end}
            \ENDFOR 
        \ENDIF 
    \ENDFOR 
\ENDFOR

\end{algorithmic}
\end{algorithm}

\noindent\textbf{Mixed-integer linear program formulation and solving.} Given a CMTG $\mathbf{C}$, we find a set of task skeletons that specify which robot will move which object at each time step. We assume that each object will be moved at most once, i.e., we assume that the problem instances are \textit{monotone}. Given a time step limit $T$, we cast the problem of finding a task skeleton that has a minimum number of objects to be moved as a mixed-integer linear program (MIP). We encode the precedence of manipulating different objects as formal constraints in the MIP such that we can generate task skeletons that are promising to be successfully grounded. We incrementally increase the time step limit $T$. In our implementation, the maximum time step limit is a hyperparameter.

For simplicity of presentation, we slightly abuse $\mathbf{M}$ again to denote the objects in $\mathbf{C}$. We use $\mathbf{M}^* \subseteq \mathbf{M}$ to denote the objects that are intended to be moved. We slightly abuse $\mathbf{\bar{a}}$ to denote the set of partially grounded pick-and-place actions in $\mathbf{C}$. We use $E_{\mathbf{\bar{a}}} = \{(M, \bar{a})\}$ to denote the set of action edges in $\mathbf{C}$. We use $E^{pick}_B = \{(\bar{a}, M)\}$ to denote the set of block-pick edges and $E^{place}_B = \{(\bar{a}, M)\}$ to denote the set of block-place edges in $\mathbf{C}$, $E_B = E^{pick}_B \cup E^{place}_B$,  where $M \in \mathbf{M}$ and $\bar{a} \in \mathbf{\bar{a}}$. We define the binary variables $X^t_{M, \bar{a}}$ and $X^t_{\bar{a}, M}$, where $t \in [1, \dots, T], (M, \bar{a}) \in E_{\mathbf{\bar{a}}}$ and $(\bar{a}, M) \in E_B$. $X^t_{M, \bar{a}} = 1$ implies that action $\bar{a}$ is executed at time step $t'\text{ s.t. }t' \geq t$. $X^t_{\bar{a}, M} = 1$ implies that object $M$ can be considered for being moved at time step $t$ since it blocks action $\bar{a}$ which is executed at or after time step $t$. 




Our MIP model is shown in the following. The implications in constraint $(11)$ and constraint $(12)$ are compiled to linear constraints using the big-M method~\cite{griva2009linear}:
{
\scriptsize
\begin{align}
&\text{minimize} \sum\nolimits_{(M, \bar{a}) \in E_{\mathbf{\bar{a}}}} X^1_{M, \bar{a}} \nonumber \\
&X^{t}_{M, \bar{a}} \geq X^{t + 1}_{M, \bar{a}}, \forall (M, \bar{a}) \in E_{\mathbf{\bar{a}}}, t \in [1, T - 1] \\
&X^t_{M, \bar{a}} = X^t_{\bar{a}, M'}, \forall (M, \bar{a}) \in E_{\mathbf{\bar{a}}}, (\bar{a}, M') \in E_B, t \in [1, T] \\
&X^t_{M, \bar{a}'} \leq \sum\nolimits_{(\bar{a}, M) \in E_{B}} X^t_{\bar{a}, M}, \forall M \in \mathbf{M} \setminus \mathbf{M}^*, (M, \bar{a}') \in E_{\mathbf{\bar{a}}}, \nonumber \\
&~~~~~~~~~~~~~~~~~~~~~~~~~~~~~~~~~~~~~~~~~~~~~~~~~~~~~~~~~~~~~~~~~~~~~~~~ t \in [1, T] \\
&\sum\nolimits_{(M, \bar{a}) \in E_{\mathbf{\bar{a}}} \text{ s.t. } R \text{ in } \bar{a}} X^T_{M, \bar{a}} \leq 1, \forall R \in \mathbf{R} \\
&\sum\nolimits_{(M, \bar{a}) \in E_{\mathbf{\bar{a}}}} X^T_{M, \bar{a}} \geq 1 \\
&\sum\nolimits_{(M, \bar{a}) \in E_{\mathbf{\bar{a}}} \text{ s.t. } R \text{ in } \bar{a}} X^t_{M, \bar{a}} \leq 1 + \sum\nolimits_{(M, \bar{a}) \in E_{\mathbf{\bar{a}}} \text{ s.t. } R \text{ in } \bar{a}} X^{t + 1}_{M, \bar{a}}, \nonumber \\ &~~~~~~~~~~~~~~~~~~~~~~~~~~~~~~~~~~~~~~~~~~~~~~~~~~~~~~~\forall R \in \mathbf{R}, t \in [1, T - 1] \\
&\sum\nolimits_{(M, \bar{a}) \in E_{\mathbf{\bar{a}}}} X^t_{M, \bar{a}} \geq 1 + \sum\nolimits_{(M, \bar{a}) \in E_{\mathbf{\bar{a}}}} X^{t + 1}_{M, \bar{a}}, t \in [1, T - 1] \\
&\sum\nolimits_{(M, \bar{a}) \in E_{\mathbf{\bar{a}}}} X^1_{M, \bar{a}} = 1, \forall M \in \mathbf{M}^* \\
&\sum\nolimits_{(M, \bar{a}') \in E_{\mathbf{\bar{a}}}} X^1_{M, \bar{a}'} \geq X^1_{\bar{a}, M},  \forall (\bar{a}, M) \in E_B\\
&\sum\nolimits_{(M, \bar{a}) \in E_{\mathbf{\bar{a}}}} X^1_{M, \bar{a}} \leq 1, \forall M \in \mathbf{M} \\
& X^1_{\bar{a}, M} = 1 \implies \sum\nolimits_{t \in [1, \dots, T]} X^t_{\bar{a}, M} \geq \nonumber \\ &(\sum\nolimits_{(M, \bar{a}') \in E_{\mathbf{\bar{a}}}} \sum\nolimits_{t \in [1, \dots, T]} X^t_{M, \bar{a}'}) + 1, \forall (\bar{a}, M) \in E^{pick}_B \\
& X^1_{\bar{a}, M} = 1 \implies \sum\nolimits_{t \in [1, \dots, T]} X^t_{\bar{a}, M} \geq \nonumber \\ &(\sum\nolimits_{(M, \bar{a}') \in E_{\mathbf{\bar{a}}}} \sum\nolimits_{t \in [1, \dots, T]} X^t_{M, \bar{a}'}), \forall (\bar{a}, M) \in E^{place}_B
\end{align}
\normalsize
}


Constraint $(1)$ enforces that $X^t_{M, \bar{a}}$ indicates whether we have selected $\bar{a}$ at or after time step $t$. Constraint $(2)$ enforces that, if an action is selected, then the objects that obstruct it are also moved. Constraint $(3)$ enforces that, besides the objects in $\mathbf{M}^*$, we only move objects that obstruct the actions we have selected. Constraints $(4-7)$ enforce that, at each time step, we select at least one action, while each robot executes at most one action. Constraint $(8)$ enforces that the objects in $\mathbf{M}^*$ are moved. Constraint $(9)$ enforces that all obstacles for the selected actions are moved, while constraint $(10)$ enforces that each object is moved only once. Constraint $(11)$ enforces that each object is moved after the obstacles for its pick action have been moved. Constraint $(12)$ enforces that each object is moved after the obstacles for its place action have been moved. The objective function represents the number of moved objects.



From a MIP solution, we construct a task skeleton which is grounded later. Moreover, we want to construct multiple task skeletons since some task skeletons may be impossible to ground. Every time we obtain a solution, we add a constraint to the MIP model to enforce that we find a different solution from the existing ones until we collect enough task skeletons~\cite{10.1007/978-3-540-72792-7_22}. In our implementation, the maximum number of task skeletons is a hyperparameter that varies for different problem instances.

\subsection{Key Component 2: Task-Skeleton Grounding}\label{sec:inst}
The second key component in the search phase (Sec.~\ref{sec:tree_search}) is to ground the task skeletons, i.e., to find the object placements and motion trajectories for the partially grounded pick-and-place actions. We use a reverse search algorithm inspired by~\cite{4209604} since forward search for continuous parameters of long-horizon task skeletons without any guidance is very challenging~\cite{Kim2019}. The insight behind the reverse search strategy is to use the grounded future joint actions as the artificial constraints to guide the grounding for the present time step.

The input to this component is a task skeleton $\bar{\mathbf{S}}$ of $T$ time steps and a sequence $\mathbf{S}_{fut}$ of future grounded joint actions. We denote the volume of work space occupied by grounded joint actions $\mathbf{S}_{fut}$ as $V_{fut}$. We denote the set of movable objects that will be moved by grounded joint actions $\mathbf{S}_{fut}$ as $\mathbf{M}_{fut}$. We denote the set of movable objects that will not be moved by task skeleton $\bar{\mathbf{S}}$ and grounded joint actions $\mathbf{S}_{fut}$ as $\mathbf{M}_{out}$. For time step $t \in [1, \dots, T]$, we denote the set of objects that are planned to be moved as $\mathbf{M}^t$ and the set of robots that are planned to move them as $\mathbf{R}^t$. Recall that we denote the goal specification and the set of movable objects as $\mathcal{G}$ and $\mathbf{M}$, respectively.


The grounding starts at the last time step $T$. For time step $t$, we first sample placements for objects $\mathbf{M}^t$ that are collision-free with respect to objects $\mathbf{M}_{out} \cup \mathbf{M}_{fut}$, fixed objects $\mathbf{F}$ and volume $V_{fut}$. The sampled placements should not collide with volume $V_{fut}$, because, otherwise, they will prevent the execution of future grounded joint actions that occupy $V_{fut}$. 

Given the placements, we plan pick trajectories and place trajectories for objects $\mathbf{M}^t$ and robots $\mathbf{R}^t$ that are collision-free with respect to objects $\mathbf{F} \cup \mathbf{M}_{fut} \cup \mathbf{M}_{out}$. We note that, in addition to the fixed objects $\mathbf{F}$ and the objects $\mathbf{M}_{out}$, the planned trajectories should not collide with the objects $\mathbf{M}_{fut}$ that are moved in future grounded joint actions.

Since we may move multiple robots and objects concurrently, we do not allow collisions between the robots, collisions between the moved objects and collisions between a robot and a moved object that is not intended to be manipulated by that robot. If we succeed in grounding the joint action at time step $t$, then we expand volume $V_{fut}$ with the volume occupied by the newly planned robot and object trajectories, expand the set $\mathbf{M}_{fut}$ with the moved objects $\mathbf{M}^t$ and expand the grounded joint actions $\mathbf{S}_{fut}$ with the newly grounded joint action. We then start to ground the joint action at time step $t - 1$. If we succeed in grounding the joint actions at every time step, we return an executable task-and-motion plan $\mathbf{S}^* = \mathbf{S}_{fut}$. However, if we fail at grounding the joint action at time step $t$, we relax the collision constraints by allowing the sampled placements and trajectories to collide with the objects $\mathbf{M}_{out}$ since we can generate new skeletons to move them later. If we succeed after relaxing the constraints, then we terminate the grounding and return the sequence of the grounded joint actions $\mathbf{S}' = \mathbf{S}_{fut}$ and a set of objects $\mathbf{M}^*$. The set of objects $\mathbf{M}^*$ consists of the objects that are named in the goal specification $\mathcal{G}$ but have not yet been moved and the movable objects in the environment that occlude the grounded joint actions $\mathbf{S}'$. During the search process (Sec.~\ref{sec:tree_search}), the returned $\mathbf{S}'$ and $\mathbf{M}^*$ are then used as input to the first key component (Sec.~\ref{sec:approx}) to generate new task skeletons. If, after relaxing the collision constraints, we still cannot find feasible placements and paths, then we simply return failure.

\section{EXPERIMENTS}\label{sec:exp}

We empirically evaluate our framework in two challenging domains and show that it can generate high-quality collaborative task-and-motion plans more efficiently than two baselines.

\subsection{Baselines}

We compare our framework with two state-of-the-art TAMP frameworks. We provide both baseline planners with information about the reachable regions of each robot.

Ap1 is a multi-robot extension of the RSC algorithm~\cite{4209604} by assuming that the robots form a single composite robot. The action space includes all possible combinations of the single-robot actions and collaboration actions.

Ap2 is a general MR-TAMP framework~\cite{9636119} that is efficient in searching for promising task plans based on the constraints incurred during motion planning. We implemented the planner in a way such that geometric constraints can be utilized efficiently, e.g., the planner can identify that it needs to move the blocking objects away before it can manipulate the blocked objects.

\begin{table*}[ht]\centering
\caption{Comparison of the proposed method with two baseline methods in the two benchmark domains regarding the success rate, planning time, makespan and motion cost. The numbers in the names of the problem instances indicate the numbers of the goal objects and the movable objects besides the goal objects. In PA5, PA7 and PA10, each problem instance has $3$ goal objects and $2$ robots. We omit the planning time and solution quality results for Ap2 on PA10 because its success rate is significantly lower than those of the other two methods.}
\resizebox{1.\linewidth}{!}{
\begin{tabular}{ccccccccccccc}
    \hline
    \\ [-1em]
    Problem Instance & \multicolumn{3}{c}{Success rate \%} & \multicolumn{3}{c}{Planning time (s)} & \multicolumn{3}{c}{Makespan} & \multicolumn{3}{c}{Motion cost} \\
    \\ [-1em]
    \cmidrule(lr){2-4} \cmidrule(lr){5-7} \cmidrule(lr){8-10} \cmidrule(lr){11-13}
    & Ap1 & Ap2 & Ours & Ap1 & Ap2 & Ours  & Ap1 & Ap2 & Ours & Ap1 & Ap2 & Ours \\
    \\ [-1em]
    \hline
    \\ [-1em]
    PA5 & \textbf{100.0} & 80.0 & \textbf{100.0} & 5.6 ($\pm$1.3) & 6.1 ($\pm$2.1) & \textbf{2.4 ($\pm$0.2)}  & 3.0 ($\pm$0.2) & 2.9 ($\pm$0.2) & \textbf{2.8 ($\pm$0.2)}  & 3.8 ($\pm$0.2) & \textbf{3.6 ($\pm$0.2)} & \textbf{3.6($\pm$0.2)} \\
    \\ [-1em]
    PA7 & 80.0 & 70.0 & \textbf{100.0} &  39.8 ($\pm$12.8) & 10.5 ($\pm$2.9) & \textbf{4.0 ($\pm$0.9)} & 3.7 ($\pm$0.3) & \textbf{3.0 ($\pm$0.3)} & 3.1 ($\pm$0.2) & 4.8 ($\pm$0.3) & 4.3 ($\pm$0.2) & \textbf{4.1 ($\pm$0.2)}\\
    \\ [-1em]
    PA10 & 55.0 & 40.0 & \textbf{90.0} & 129.2 ($\pm$58.2) & N/A & \textbf{19.6 ($\pm$6.1)}  & 4.6 ($\pm$0.6) & N/A & \textbf{4.2 ($\pm$0.3)} & 5.6 ($\pm$0.6) & N/A & \textbf{5.2 ($\pm$0.4)}\\
    \\ [-1em]
    BO8 & 85.0 & N/A & \textbf{100.0} & 246.5 ($\pm$54.2) & N/A & \textbf{182.2 ($\pm$48.3)}  & 4.8 ($\pm$0.2) & N/A & \textbf{3.4 ($\pm$0.3)} & 7.6 ($\pm$0.1) & N/A & \textbf{5.0 ($\pm$0.6)}\\

    \hline 
\end{tabular}
}
\label{tab:exp_time}
\end{table*}

\subsection{Benchmark Domains}

We evaluate the efficiency and effectiveness of our method and the two baselines in the \textbf{packaging} domain shown in Fig.~\ref{fig:best} (left) and the \textbf{box-moving} domain shown in Fig.~\ref{fig:best} (right).

\noindent\textbf{Packaging (PA):} In this domain, each problem instance includes $2$ to $4$ robots, $3$ to $5$ goal objects, $2$ to $13$ movable objects besides the goal objects, $1$ start region and $3$ goal regions. As in~\cite{doi:10.1177/02783649211038280}, we omit motion planning and simply check for collisions at the picking and placing configurations computed by inverse kinematics solvers in this domain, because collisions in this domain mainly constrain the space of feasible picking and placing configurations. We use Kinova Gen2 lightweight robotic arms. For each benchmark problem instance, we conduct $20$ trials with a timeout of $1,200$ seconds. For all methods, we also count a trial as failed, if all possible task plans have been tried.

\noindent\textbf{Box-moving (BO):} In this domain, each problem instance includes $2$ robots, $2$ goal objects, $6$ movable objects besides the goal objects, $1$ start region and $1$ goal region. We use PR2 robots. In this domain, we do not consider handover actions, because, they do not contribute significantly to generating feasible and high-quality plans for MR-GTAMP problems with mobile robots in synchronous setups. For each benchmark problem instance, we conduct $20$ trials with a timeout of $1,200$ seconds. For both methods, we also count a trial as failed, if all possible task plans have been tried. In this domain, we compare our method only with Ap1, since Ap2 is restricted to manipulators.

We use bidirectional rapidly-exploring random trees~\cite{lavalle2006planning} for motion planning and IKFast~\cite{diankov_thesis} for inverse kinematics solving. All methods share the same grasp sets, the same sets of single-robot actions, and the same sets of collaboration actions. All experiments were run on an AMD Ryzen Threadripper PRO 3995WX Processor with a memory of 64GB.

\subsection{Results}
We refer to the number of time steps as \textit{makespan} and the number of moved objects as \textit{motion cost}.

\noindent\textbf{Planning time and success rate.} Table~\ref{tab:exp_time} shows that our method outperforms both baseline methods on all problem instances with different numbers of goal objects and movable objects with respect to both the planning times and success rates. Ap1 and our method achieve higher success rates on all problem instances than Ap2 because the reverse search strategy (Sec.~\ref{sec:inst}) utilized in Ap1 and our method finds feasible object placements much more efficiently than the forward search strategy used in Ap2. Moreover, Ap2 can generate task plans that include irrelevant objects while Ap1 and our method focus on manipulating the important objects, like blocking objects for necessary manipulation or goal objects. Our method achieves higher success rates with shorter planning times than Ap1 on the difficult problem instances PA7, PA10 and BO8 because our method first generates promising task skeletons (Sec.~\ref{sec:approx}) that use the information about the collaborative manipulation capabilities of the individual robots to prune the task plan search space, which can be extremely large when there are many objects and multiple robots~\cite{9636119}. The main cause of failure of our method is running out of task skeletons which can be addressed by incrementally adding more task skeletons during the search process.

\noindent\textbf{Solution quality.} Table~\ref{tab:exp_time} shows that our method can generate high-quality task-and-motion plans with respect to the motion cost and the makespan. Our method first generates task skeletons with short makespans by incrementally increasing time step limit and with low motion costs by incorporating the motion cost into the objective function of the MIP formulation (Sec.~\ref{sec:approx}). On the other hand, our MCTS exploration strategy motivates the planner to search for high-quality plans with small numbers of moved objects. It should be noted that, although Ap2 generates plans with shorter makespans for PA7, it has lower success rates and longer planning times than our method. Also, Ap1 generates plans that move significantly more objects for PA7, PA10 and BO8 than our method because it uses a depth-first search strategy for finding feasible plans~\cite{4209604}. 

\begin{table}[ht]\centering
\caption{The results of the proposed method in domain PA regarding the success rate, planning time, makespan and motion cost. The numbers in the names of the problem instances indicate the numbers of the robots.}
\resizebox{1.\columnwidth}{!}{\begin{tabular}{ccccc}

    \hline
    \\ [-1em]
    Problem Instance & \multicolumn{1}{c}{Success rate \%} & \multicolumn{1}{c}{Planning time (s)} & \multicolumn{1}{c}{Makespan} & \multicolumn{1}{c}{Motion cost} \\
    \cmidrule(lr){2-2} \cmidrule(lr){3-3} \cmidrule(lr){4-4} \cmidrule(lr){5-5} 
    \\ [-1em]
    2 robots & 60.0 & 148.4 ($\pm$36.8) & 6.1 ($\pm$0.4) & 8.9 ($\pm$0.4)\\
    \\ [-1em]
    3 robots & 80.0 & 99.0 ($\pm$48.6) & 4.9 ($\pm$0.3) & 8.2 ($\pm$0.5) \\
    \\ [-1em]
    4 robots & 85.0 & 109.1 ($\pm$33.6) & 4.7 ($\pm$0.3) & 8.2 ($\pm$0.4) \\
    \\[-1em]
    \hline 
\end{tabular}}
\label{tab:scale}
\end{table}
\noindent\textbf{Scalability evaluation.} We evaluate the scalability of our method in the PA domain with $18$ movable objects, including $5$ goal objects, and $2$ to $4$ robots. Table~\ref{tab:scale} shows that our method can solve these large problem instances. Moreover, for problem instances with $3$ and $4$ robots, it achieves higher success rates, shorter makespans and lower motion costs compared to the problem instances with $2$ robots. This shows that our method can generate intelligent collaboration strategies for multiple robots.

\section{CONCLUSION}
In this paper, we presented a framework for MR-GTAMP problems by proposing a novel MIP formulation to utilize information about the collaborative manipulation capabilities of the individual robots to generate promising task skeletons for guiding the planning search. We proposed an efficient task-skeleton grounding algorithm inspired by the previous work on MAMO~\cite{4209604}. The proposed components are integrated via a Monte-Carlo Tree Search exploration strategy that searches for high-quality task-and-motion plans. We showed that our framework outperforms two baselines on two challenging MR-GTAMP problems with respect to the planning time and success rates, can generate high-quality plans with respect to the resulting plan length and the number of objects moved, and can scale up to large problem instances.

While we have assumed full observability of the scene, we plan to account for sensing limitations in the future~\cite{nikolaidis2009optimal,7451762}. Future work also includes using learning to improve the planning efficiency~\cite{doi:10.1177/02783649211038280} and extending the developed techniques to more general MR-TAMP problems~\cite{pmlr-v100-kim20a} and more diverse environments~\cite{FontaineB-RSS-21,Zhang_Fontaine_Hoover_Togelius_Dilkina_Nikolaidis_2020}.






\section*{Acknowledgements}
This work was supported by the National Science Foundation NRI \# 2024936 and the Alpha Foundation \# AFC820-68.




\small
\bibliographystyle{IEEEtran}
\bibliography{literature}

\vfill

\end{document}